\documentclass[letterpaper, 10 pt, journal, twoside]{IEEEtran}

\usepackage{graphicx} 
\usepackage{epsfig} 
\usepackage{mathptmx} 
\usepackage{times} 
\usepackage{amsmath} 
\usepackage{amssymb}  
\usepackage[capitalize]{cleveref}
\usepackage{array}
\usepackage{multirow}
\usepackage{tikz}
\usepackage{url}
\usepackage[T1]{fontenc}
\usepackage{comment}
\usepackage{todonotes}
\usepackage{enumitem}

\title{AIVIO: Closed-loop, Object-relative Navigation of UAVs with AI-aided Visual Inertial Odometry}
\author{Thomas Jantos$^{1}$, Martin Scheiber$^{1}$, Christian Brommer$^{1}$, Eren Allak$^{1}$, Stephan Weiss$^{1}$ and Jan Steinbrener$^{1}$
\thanks{Manuscript received: July, 4, 2024; Revised September, 11, 2024; Accepted October, 4, 2024.}
\thanks{This paper was recommended for publication by Editor Pascal Vasseur upon evaluation of the Associate Editor and Reviewers' comments. This work was supported by the Federal Ministry for Climate Action, Environment, Energy, Mobility, Innovation and Technology (BMK) under the grant agreement 881082 (MUKISANO) and the Austrian Science Fund (FWF): TAI 183.}
\thanks{$^{1}$The authors are with the Control of Networked Systems Group, University of Klagenfurt, 9020 Klagenfurt am Wörthersee, Austria {\tt\footnotesize \{name.surname\}@ieee.org}}%
\thanks{{\textbf{Pre-print version, accepted Oct/2024, DOI follows ASAP~\copyright IEEE.}}}
}

\begin{document}
\maketitle
\begin{abstract}
Object-relative mobile robot navigation is essential for a variety of tasks, e.g. autonomous critical infrastructure inspection, but requires the capability to extract semantic information about the objects of interest from raw sensory data. While deep learning-based (DL) methods excel at inferring semantic object information from images, such as class and relative 6 degree of freedom (6-DoF) pose, they are computationally demanding and thus often not suitable for payload constrained mobile robots. In this letter we present a real-time capable unmanned aerial vehicle (UAV) system for object-relative, closed-loop navigation with a minimal sensor configuration consisting of an inertial measurement unit (IMU) and RGB camera. Utilizing a DL-based object pose estimator, solely trained on synthetic data and optimized for companion board deployment, the object-relative pose measurements are fused with the IMU data to perform object-relative localization. We conduct multiple real-world experiments to validate the performance of our system \textcolor{black}{for the challenging use case of power pole inspection}. An example closed-loop flight is presented in the supplementary video.
\end{abstract}



\vspace{-0.4cm}
\section{Introduction}

Semantic navigation describes the robots ability to maneuver and perform tasks based on contextual information from its environment. Object-relative localization, one of the core disciplines of semantic navigation, is essential for a variety of mobile robot applications such as infrastructure inspection and object tracking. Both tasks require the mobile robot to extract semantic information from its onboard sensors, i.e., detecting objects of interest and estimating their 6 degree of freedom (6-DoF) pose. In recent years, unmanned aerial vehicles (UAVs) gained a lot of popularity for mobile robot applications due to their ability to freely move in 3D space. While this allows for the deployment in  a variety of scenarios, it comes at the cost of constraints, i.e. weight, size, amount of sensors and computational power. Moreover, UAVs require constant control inputs based on reliable and rapid state estimates to perform even simple maneuvers such as hovering. Therefore, to guarantee fully autonomous, closed-loop navigation, \textcolor{black}{i.e. applying control inputs based on the current state estimate to follow a trajectory},  algorithms and sensor processing need to run in real-time onboard the UAV, which requires efficient and optimized methods.
\looseness=-1

Fusing information from multiple sources, i.e. sensors, is necessary to guarantee reliable and accurate state estimates. Inertial measurement units (IMUs) provide UAVs with linear acceleration and angular velocity measurements at a high frequency. Solely relying on IMUs for state estimation is not feasible as this would lead to drift and thus UAVs often rely on GNSS for localization and navigation in outdoor scenarios. The need for GNSS-free state estimation arises due to possible signal deterioration caused by large structures or multipathing. Infrastructure inspection applications require precise, object-relative navigation capabilities down to the centimeter range to reliably position the UAV for accurate analysis. This cannot be achieved with \textcolor{black}{the measurement accuracy of} GNSS. While classical sensor fusion considering other sensor modalities, e.g. cameras or radar, allow for robust and precise state estimation, i.e. Visual-Inertial Odometry (VIO) \cite{fornasier2024msceqf} and Radar-Inertial Odometry (RIO) \cite{michalczyk2023radar} respectively, they do not provide any semantic information about the current scene.
\looseness=-1

\begin{figure}
    \centering
    \includegraphics[width=1.\columnwidth]{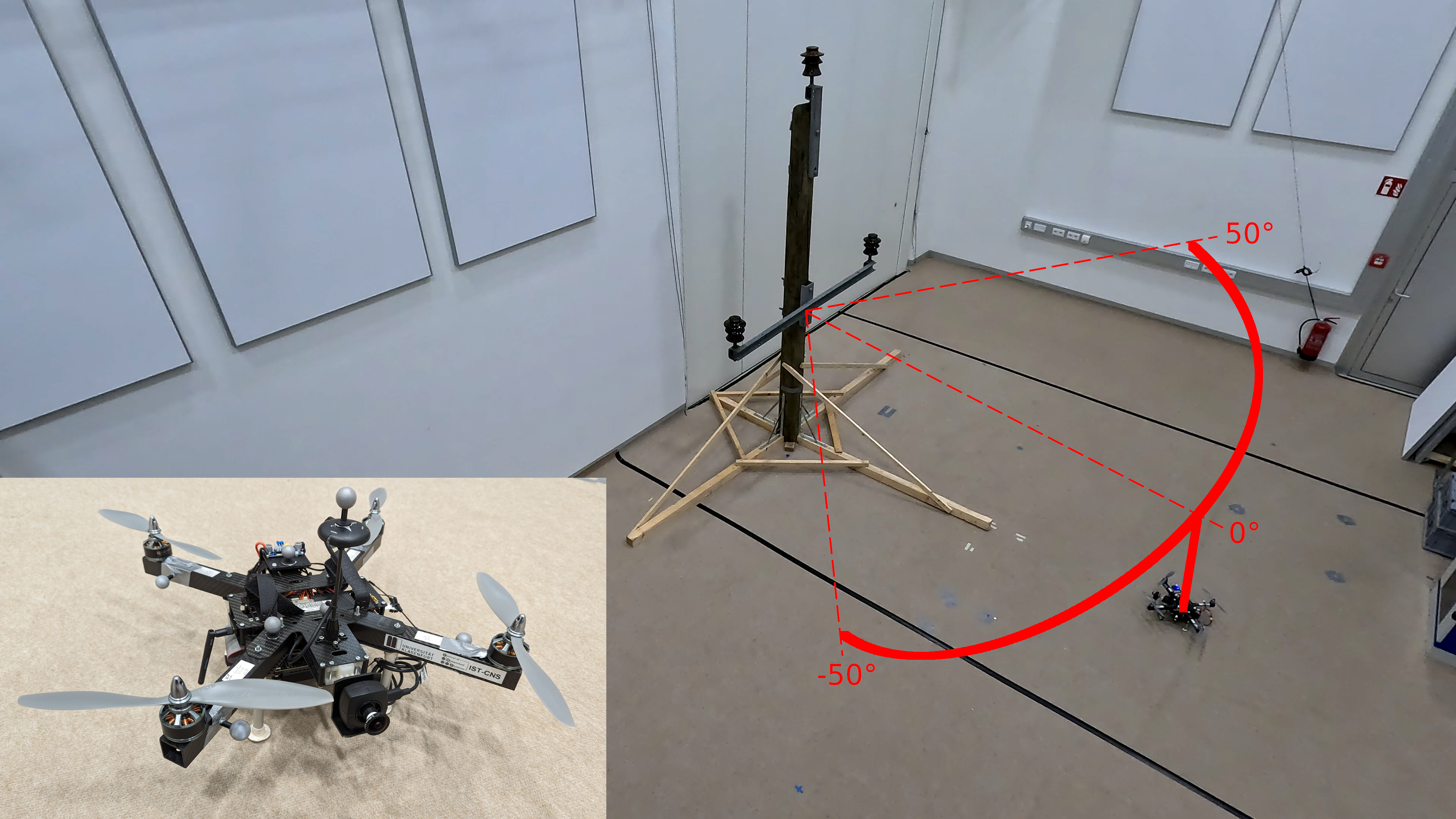}
    \caption{Experimental setup of a mock-up power pole with three insulators in our research laboratory. Moreover, we visualize a trajectory in red representing an inspection flight. \textit{Bottom left}: Our mobile robot platform of choice, a quadcopter equipped with a PX4, an RGB camera and an NVIDIA Jetson.}
    \label{fig:title_picture}
    \vspace{-0.6cm}
\end{figure}

Deep learning (DL) excels at extracting semantic information from raw sensory data, particularly for image-based perception tasks including object detection, classification and 6-DoF object pose estimation. The latter provides metric information about objects present in the scene which can also be used for 
object-relative localization, building the foundation for semantic navigation in inspection tasks. DL-based methods are usually computationally intensive, and the real-time capability necessary for closed-loop navigation is often not given. This problem is even more pronounced in the case of deployment on edge devices. We utilize optimization techniques and a computationally light DL-based object pose estimator to address this issue.

In this letter, we present a real-time capable system for object-relative UAV flights. We consider a minimal sensor configuration consisting of a single RGB camera and an IMU, allowing our approach to be used on a wide variety of resource constrained mobile robots. We investigate our approach in the context of power pole infrastructure inspection with insulators as our objects of interest. Infrastructure inspection is crucial to ensure safety, prevent failures, and maintain the reliability and longevity of essential public systems and services. The UAV performs an object-relative inspection flight by \textcolor{black}{solely} relying on \textcolor{black}{IMU measurements and} relative-pose measurements to the insulators\textcolor{black}{,} without any prior knowledge about their location \textcolor{black}{or additional sensors, e.g. GNSS}. Our experimental setup, the inspection trajectory and our UAV are presented in \cref{fig:title_picture}. Training DL methods is resource and data intensive. Especially the latter poses a problem if suitable datasets are not available. While there exist methods for the semi-automatized annotation of 6-DoF pose datasets \cite{suchi20233ddat}, they either require a motion capture system (MoCap) or initial human input. On the other hand, simulation software allows for the efficient and automated generation of large datasets that also capture the environmental diversity present in infrastructure inspection tasks.  
It is important to note that, even though we focus solely on power poles throughout this letter, our approach is applicable to many object-relative localization and navigation scenarios, in which the 6-DoF pose of the objects of interest is well-defined. Our contributions are the following:

\begin{itemize}[leftmargin=*,noitemsep,topsep=0pt,parsep=0pt,partopsep=0pt]
    \item UAV system for object-relative, closed-loop navigation.
    \item Onboard accurate 6-DoF object-relative localization by \mbox{fusing} estimated DL-based 6-DoF object-relative poses from images with IMU measurements. 
    \item Illustration of the Sim2Real transfer capabilities of the \\6-DoF object-relative pose estimator solely trained on \mbox{simulated} data. 
    \item Development of a camera-agnostic approach by performing a homography to map between different camera parameters.
    \item Validation of the proposed system by conducting several real world experiments to show the accuracy and performance.
\end{itemize}

\noindent The remainder of this letter is structured as follows. In \cref{sec:related_work}, we summarize the related work regarding object-relative localization in mobile robotics. In \cref{sec:method}, we present our system including the hardware and software setup, the 6-DoF object pose estimator and state estimator. In \cref{sec:experiments}, the experiments and the corresponding results are discussed. Finally, the letter is concluded in \cref{sec:concluison}.
\looseness=-1
 
\vspace{-0.3cm}
\section{Related Work} \label{sec:related_work}

In mobile robotics, navigation requires accurate state estimation and it typically involves merging IMU data with one or more sensors like GNSS. While GNSS offers global position information, but lacks 3D orientation data, it still can be used to estimate the attitude of a robot by combining it with IMU measurements \cite{kornfeld1998single}. In the absence of GNSS signals, Visual-Inertial Odometry (VIO), which combines a monocular camera with IMU data, can determine the robot's pose by triangulating the camera's position using image features and estimating the scale factor with inertial data \cite{weiss2011monocular}. Nonetheless, the vision pose is prone to drift in position over time. Integrating multiple sensors enhances the accuracy, reliability and robustness of the state estimate. Two primary approaches for sensor fusion exist: \textcolor{black}{recursive, filter-based} methods and optimization-based techniques. Although optimization-based methods can provide more precise state estimates, they are computationally intensive due to optimizing across multiple sensor measurements \cite{sola2022wolf}. In contrast, filter-based methods like Extended Kalman Filters (EKF) are computationally efficient, making them suitable for mobile robotics applications \cite{brommer2020mars}.

Classical VIO works on raw image features contained in the image that do not provide any information about the objects of interest present in the environment and thus is not suited for object-relative state estimation and localization. However, methods for estimating the relative 6-DoF object pose from images can be employed as a pose sensor within a state estimation framework \cite{jantos2023ai}. DL-based methods for monocular 6-DoF pose estimation of known objects differ in the number of viewpoints, input modalities, and 3D object model usage. While observing the object from multiple points of view introduces constraints on its pose \cite{labbe2020cosypose}, utilizing depth maps as an additional input provides the network with preliminary object distances \cite{hodavn2015detection}. The availability of 3D object models allows for the iterative refinement of an initial guess \cite{li2018deepim} or the matching of keypoints \cite{wang2021gdr}. Even though this additional information benefits the estimation process, they often require either additional sensors or are computationally more complex compared to their single-view, RGB-only counterparts. The 6-DoF multi-object pose estimation framework PoET \cite{jantos2023poet}, which we recently presented, only requires a single RGB image as input and does not require any extra information during training or inference. It achieves state-of-the-art performance on benchmark datasets.
\looseness=-1

Earlier research on vision-based, object-relative state estimation for UAVs utilizes geometrical features of the object to determine its pose. Thomas et al. \cite{thomas2015visual} presented an approach for localization relative to cylindrical objects of interest by assuming the radius to be known. Similarly, Loianno et al. \cite{loianno2018localization} used the parametric description of ellipses to detect objects of interest in the image and determine their pose based on their visual appearance, e.g., size, and camera parameters. Meanwhile, M{\'a}th{\'e} et al. \cite{mathe2016vision} utilized classical methods to determine 6-DoF object poses for object-relative pose estimation, while also investigating the possibility of using machine learning to detect the presence of objects in the image.

An alternative to object-relative state estimation based on the geometric features of objects is to attach markers to them. Due to the predefined characteristics of these markers, their visual appearance can be used to directly derive their relative pose to the robot. For example, in \cite{do2023vision} a UAV's pose relative to a perching target was estimated by constantly fusing relative pose measurements from two ArUco markers in a Kalman filter. VI-RPE \cite{teixeira2018vi} showcased the ability to navigate a UAV relative to a target UAV equipped with a LED marker configuration. The authors used an RGB camera to determine the relative pose of the UAV and fused this measurement with IMU data for object-relative state estimation. Both \cite{shen2023aeronet} and \cite{patruno2019vision} equip a landing platform with a human-made mark for autonomous landing. The latter approach fuses the relative pose between a UAV and a landing platform with GNSS and IMU data in an EKF for accurate landing.

In our previous work \cite{jantos2023ai}, we presented the combination of DL-based object-relative pose measurements with IMU measurements for mobile robot 6-DoF multi-object-relative state estimation. We present here an efficient implementation of our previous work, trained in simulation for a realistic inspection scenario, and deployed on a real UAV for closed-loop semantic navigation. We implemented a sensor switching mechanism to allow for seamless transition between localization with a global position sensor and our DL-based, multi-object-relative pose sensor. This enables the UAV to approach the object of interest using a global positioning system and then switch to object-relative navigation once the object of interest is in the proximity and field of view of the UAV. \textcolor{black}{For object-relative navigation,} our approach does not require human-made markers, prior knowledge about the \textcolor{black}{object} configuration \textcolor{black}{or information from a global positioning sensor.} 
\vspace{-0.3cm}


\begin{figure}
    \centering
    \includegraphics[width=1.0\columnwidth]{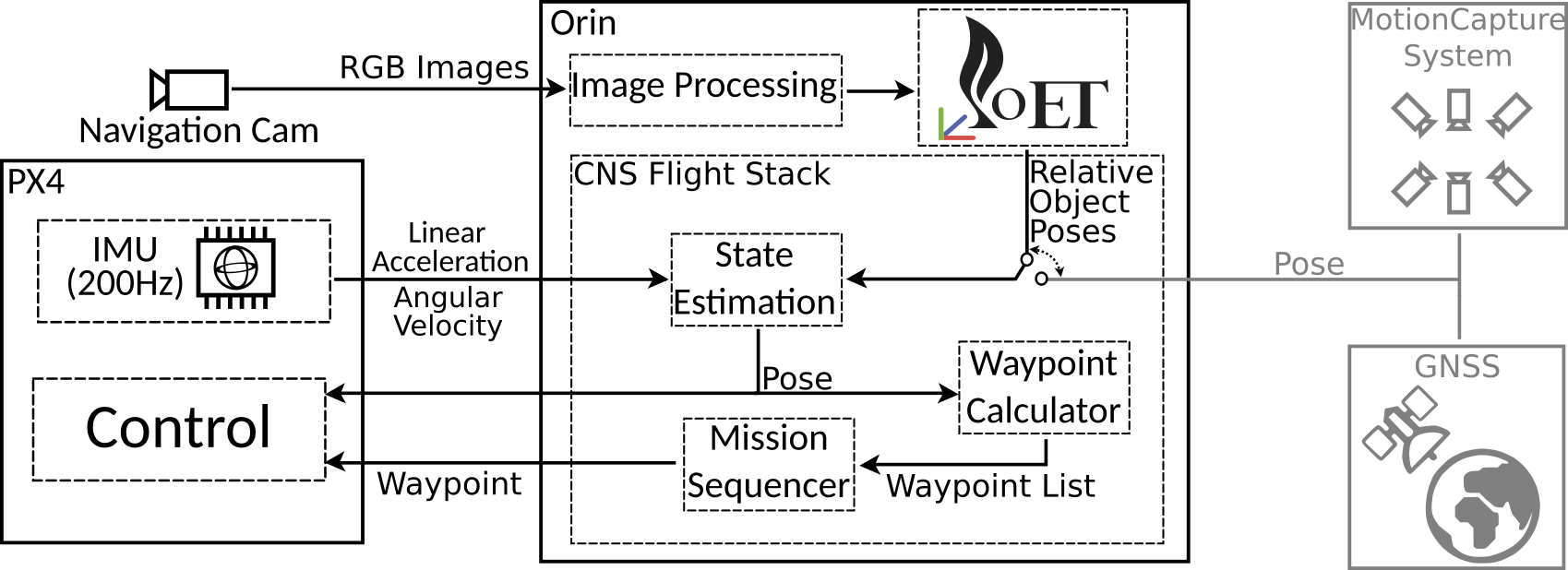}
    \caption{Schematic overview of the hardware and software components and their interaction. The CNS Flight Stack \cite{cns_flightstack22} is responsible for handling high-level autonomy and communication with the PX4. We utilize PoET \cite{jantos2023poet}, a DL-based 6-DoF object pose estimator, to predict relative object pose measurements given an input image from the RGB navigation camera. These measurements are fused with IMU data in a state estimator for object-relative localization. Moreover, the autonomy and state estimator allow for switching between a global pose sensor and our object-relative landmark sensor and to calculate trajectory waypoints based on the current estimate.}
    \label{fig:hw_sw_overview}
    \vspace{-0.5cm}
\end{figure}

\section{Method} \label{sec:method}

In this section, we present our approach. First, we explain the notation that we use for the transformation of coordinate frames. Second, the hardware setup is presented. Third, we describe the deep-learning pose estimation framework, its training and the deployment to the real hardware. Fourth, the state estimation framework is introduced and the integration with the DL-based relative pose sensor is outlined.
\vspace{-0.45cm}
\subsection{Notation} \label{subsec:notation}

\textcolor{black}{Given two coordinate frames $A$ and $B$, the transformation of frame $B$ with respect to frame $A$ is defined by the translation ${\mathbf{p}}_{\scriptscriptstyle AB}$ and rotation ${\mathbf{R}}_{\scriptscriptstyle AB}$.} $\mathbf{I}_3$ and $\mathbf{0}_3$ refer to the identity and the null matrix in $\mathbb{R}^{3x3}$, respectively. Alternatively, a rotation is expressed by a quaternion  ${\mathbf{q}}_{\scriptscriptstyle AB} = [\mathbf{q_v}~q_w]^T = [q_x ~ q_y ~ q_z ~ q_w]^T$. The quaternion multiplication is represented by $\otimes$.

\vspace{-0.3cm}
\subsection{Hardware \& Software Setup}
\label{subsec:hw_sw_setup}

Our hardware setup, as depicted in the bottom left corner of \cref{fig:title_picture}, consists of a TWINs Science Copter\footnote{https://www.twins.co.at/en/en-twinfold-science/} platform equipped with a Pixhawk PX4 autopilot, an IDS U3-3276LE-C-HQ RGB camera with a 2 MP, 2.95~mm focal length lens (angle of view: D:178$^\circ$, H:138$^\circ$, V:104$^\circ$), and an NVIDIA Jetson Orin AGX 64GB DevKit serving as the onboard computer. The IMU of the PX4 is used as the propagation sensor in our EKF and provides measurements at 200Hz. The RGB camera captures the raw images at a resolution of 1280x960 with up to 50~fps. During the launch of the camera, automatic white balance and gain adjustment is performed once, while automatic exposure control is turned off. The Orin is running on JetPack 5.1.2. For deployment of the DL networks, TensorRT 8.5.2, CUDA 11.4, and PyTorch 2.0 is used. The PX4 is connected to \textcolor{black}{pins 6 (Ground), 9 (UART1\_TX) and 10 (UART2\_RX) of} the 40 head pin of the Orin and communicates over UART and the RGB camera is connected with USB3 to the Orin. We run the CNS Flight Stack \cite{cns_flightstack22} on our onboard computer for high-level autonomy, control, mission management, safety monitoring, and data recording. The CNS Flight Stack's modularity and ROS interfaces allow us to seamlessly integrate our sensors and state estimator into the whole stack to conduct autonomous UAV flights.
An overview of the hardware and software components and their interaction is depicted in \cref{fig:hw_sw_overview}.
\looseness=-1

\vspace{-0.3cm}
\subsection{Online DL-based 6-DoF Relative Pose Estimation}
\label{subsec:method_poet}

\begin{table}
    \centering \caption{DL model comparison for different sizes, optimization and floating point precision on the synthetic test dataset.}
\scalebox{0.8}{
\begin{tabular}{l|c|c|c|c|c}
Model & Avg TE {[}m{]} & Avg RE {[}°{]} & Max TE {[}m{]} & \# Not Detected & FPS\tabularnewline
\hline 
\hline 
Torch & 0.031 & 1.52 & 0.883 & 1992 & 9\tabularnewline
TRT FP32 & 0.031 & 1.83 & 0.745 & 2060 & 14\tabularnewline
TRT FP16 & 0.031 & 1.84 & 0.763 & 2065 & 33\tabularnewline
Torch Small & 0.037 & 1.60 & 0.793 & 1927 & 13\tabularnewline
TRT Small FP32 & 0.038 & 1.94 & 0.816 & 2079 & 22\tabularnewline
TRT Small FP16 & 0.039 & 1.93 & 0.798 & 2082 & 50\tabularnewline
\end{tabular}
}
    \label{tab:torch_trt_comparison}
    \vspace{-0.3cm}
\end{table}

Deployment of computationally intensive, DL-based methods on mobile robots is an ongoing challenge due to their payload limitations. Therefore, we chose our object pose estimation framework PoET \cite{jantos2023poet} as it only uses RGB images and does not rely on depth information, removing the need for additional sensors, or 3D object models, thus reducing computational complexity. Given a single input image, PoET detects all objects of interest and predicts the relative 6-DoF pose between the camera and the objects.

\begin{figure}[t]
    \centering
    \includegraphics[width=0.45\columnwidth]{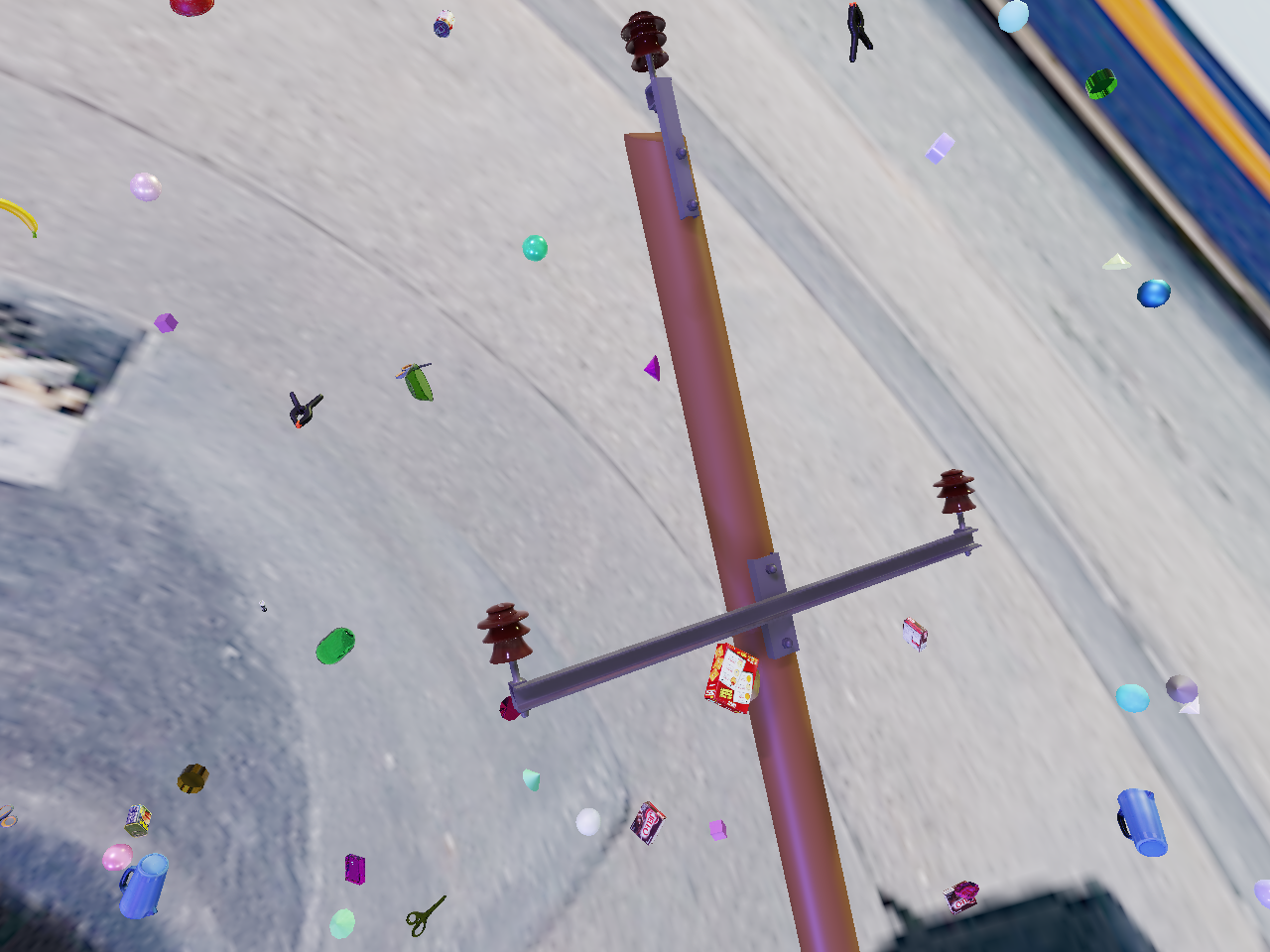}
    \includegraphics[width=0.45\columnwidth]{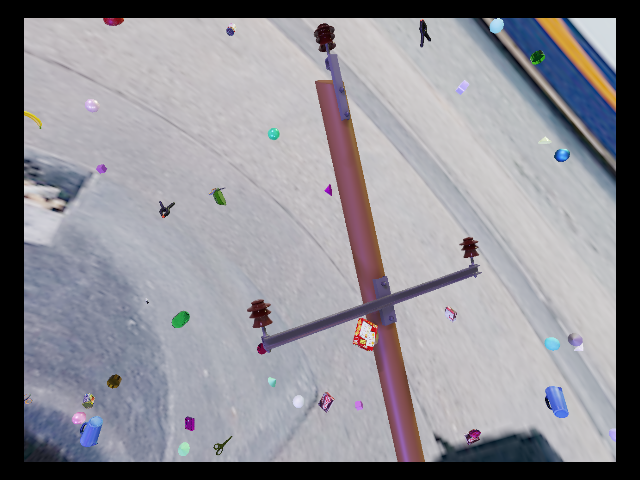}
    \caption{Visualization of the homography mapping between two different camera parameters. Note, the original sizes of the images are 1280 x 960 and 640x480, respectively for the left and right image.}
    \label{fig:homography}
    \vspace{-0.5cm}
\end{figure}

To train our DL-based object-relative pose estimator we make use of the possibility of efficiently generating images of objects of interest with simulation software. By relying solely on simulated data, the here presented approach can be easily adapted to any kind of inspection task without the need of exhaustive real-world data collection and annotation. 
We utilize NVIDIA Omniverse IsaacSim\footnote{https://developer.nvidia.com/isaac/sim} to simulate images of power poles and automatically annotate the relative 6-DoF pose of the insulators, as they are our objects of interest for the object-relative navigation. The individual components of the power pole are modeled in Blender and loaded into Isaac Sim during data generation. Besides randomized relative 6-DoF poses and the automatic annotation of the generated images, IsaacSim allows the addition of randomly moving distractor objects and a variety of randomization such as environment, background, lighting conditions, the material of objects, and the geometrical properties of the power pole in order to prevent overfitting to specific scenarios and geometric properties of the power pole. We generate 100,000 images for training and 20,000 images for validation with varying the distance to the power pole between 2.5m and 4.5m. The chosen resolution is 640x480. An example image can be seen on the left in \cref{fig:homography}.

For PoET we chose Scaled-YOLOv4 \cite{wang2021scaledyolo} as our backbone object detection network, and our base-model transformer consists of five encoder and decoder layers with 16 attention heads (\texttt{Torch}). PoET is trained class agnostic for 50 epochs and achieves an average translation and rotation error on the test set of 3.1~cm and $1.52^{\circ}$, respectively. \textcolor{black}{The training takes about 60h on a single NVIDIA GeForce RTX 3090.} We also trained a smaller network with only three encoder/decoder layers and 8 attention heads (\texttt{Torch Small}). It is important to note that even though the insulators have rotational symmetries around their z-axis, PoET can accurately estimate the relative orientation, despite being only provided the 6-DoF pose of the insulators during training. As PoET consists of a multi-head attention transformer it is able to incorporate global image context information into the pose estimation process. Hence, PoET learns the asymmetric geometry of our power pole and utilizes this information to estimate the correct orientation of the insulators.
\looseness=-1

Using TensorRT (\texttt{TRT}) we optimize both models for deployment on the NVIDIA Jetson AGX Orin either with full- (\texttt{FP32}) or half-precision (\texttt{FP16}). In \cref{tab:torch_trt_comparison}, we compare the six models with each other evaluated on the synthetic validation dataset consisting of 20,000 images with 54,104 objects in total. The comparison metrics are the translation error (\texttt{TE}) and rotation error (\texttt{RE}), given by the Euclidean and geodesic distance respectively. Moreover, we compare the models in terms of the number of missed detections and frames per second.
\looseness=-1

\begin{figure}[t]
    \centering
    \includegraphics[width = 0.45\columnwidth]{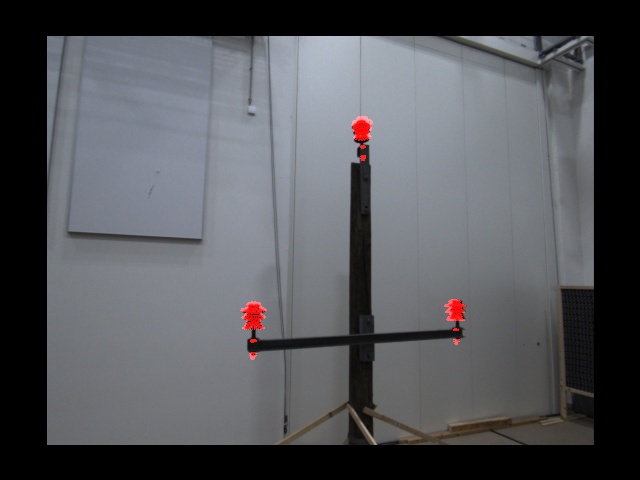}
    \includegraphics[width=0.45\columnwidth]{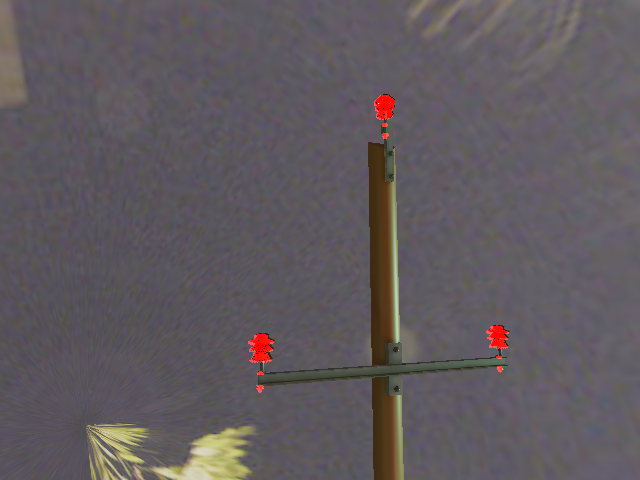}
    \caption{Reprojection of the 6-DoF object poses predicted by PoET (\texttt{TRT Small FP16}) for our real-world scenario (\texttt{left}) and a synthetic image (\texttt{right}). Note that the black border for the real image is due to the homography mapping between different camera parameters.}
    \label{fig:example_detection}
    \vspace{-0.5cm}
\end{figure}

We observe that the smaller networks have a slightly higher average error than their counterparts. Furthermore, the optimization with TensorRT has an influence on the performance of the network with respect to the average error. The lower maximum translation error for the optimized models is due to higher number of undetected objects. It appears that the TRT models are not able to detect the insulators in some difficult cases, which might result in wrong predictions and thus high errors. On the other hand, hardware specific optimization allows for higher throughput of images and thus reduced computational load. Therefore, \texttt{TRT Small FP16} is our PoET model of choice for the real-world experiments \textcolor{black}{as it has the lowest processing time while still achieving satisfying results}. It minimizes the computational load with acceptable impact on the accuracy of the prediction. In \cref{fig:example_detection}, the predictions of the chosen model are visualized for our real world scenario as well as for a synthetic image by reprojecting the 3D object model based on the predicted relative 6-DoF object poses. As can be seen, there is a high correspondence between reprojected insulators and the images, emphasizing the applicability of PoET, solely trained on synthetic data, to real images.
\looseness=-1

The 6-DoF relative object poses predicted by PoET will be used as measurements in our state estimator, as described in the subsequent subsection. To perform closed-loop navigation, we aim to provide the state estimator measurements at a rate of at least 15~Hz to avoid long duration of IMU propagation. Even though our model of choice achieves 50~fps on the NVIDIA Jetson AGX Orin, we limit the frame rate to 15~fps for our real-world experiments. This guarantees a sufficiently high measurement update rate, while limiting the computational load on the onboard computer.

\vspace{-0.3cm}
\subsection{Image Homography for Camera-Agnostic Pose Estimation}

RGB-only 6-DoF object-relative pose estimation requires that the network is trained and evaluated on images with the same camera parameters. This is not feasible for many real-world applications as this would require an identical camera setup to be used for data collection and network deployment on a mobile robot. In order to mitigate this issue and to make the trained network compatible with any RGB camera, a homography is used to relate the images taken by the navigation camera with camera matrix $\Hat{K}$ to camera matrix $K$ that was used to generate the images in simulation for network training. The homography $H$ representing a projective transformation between a target camera matrix $K$ and a source camera matrix $\Hat{K}$ is given by
\vspace{-0.15cm}
\begin{align}
    H &= T\Hat{H}= TK\Hat{K}^{-1} \\
    T &= \begin{pmatrix}
1 & 0 & u_o \\
0 & 1 & v_o \\
0 & 0 & 1 
\end{pmatrix},
\end{align}

\noindent where $u_o$ and $v_o$ are the pixel offsets between the target image center $(u_{tc}, v_{tc})$ and the projected source image center. The projected source image center $(u_{pc}, v_{pc})$ can be calculated by projecting the source image center $(u_{sc}, v_{sc})$ with $\Hat{H}$
\begin{align}
\begin{pmatrix}
     u_{pc} \\
     v_{pc} \\
     1
    \end{pmatrix} = 
    \begin{pmatrix}
     u \\
     v \\
     s
    \end{pmatrix} = \Hat{H}
    \begin{pmatrix}
        u_{sc} \\
        v_{sc} \\
        1
    \end{pmatrix}.
\end{align}

To validate the homography approach, we generate 1.000 synthetic images with a camera intrinsic matrix differing from the training dataset by $\sim2\times$ the focal length, doubling the image size and adjusting the principle point offset. In \cref{tab:torch_trt_homography} we compare the error metrics for the six networks and in \cref{fig:homography} the result of the homography is visualized. Across all models, we observe an increased average translation and rotation error compared to models that were trained and evaluated on images with the same camera intrinsics. However, taking into consideration that the minimum distance to the power pole is 2.5~m, the average translation error is only about 2\% or 5.7~cm. In \cref{sec:experiments}, we further show that our approach is camera-agnostic by utilizing this homography approach to map images from the real navigation camera for closed-loop navigation.
\looseness=-1

\begin{table}
    \centering \caption{DL model comparison for our homography approach to map between different cameras.}
\begin{tabular}{l|c|c|c}
Model & Avg TE {[}m{]} & Avg RE {[}°{]} & Max TE {[}m{]}\tabularnewline
\hline 
\hline 
Torch & 0.042 & 2.16 & 0.903\tabularnewline
TRT FP32 & 0.044 & 2.72 & 0.609\tabularnewline
TRT FP16 & 0.044 & 2.74 & 0.594\tabularnewline
Torch Small & 0.054 & 2.21 & 0.936\tabularnewline
TRT Small FP32 & 0.056 & 2.74 & 0.925\tabularnewline
TRT Small FP16 & 0.057 & 2.72 & 0.994\tabularnewline
\end{tabular}
    \label{tab:torch_trt_homography}
    \vspace{-0.5cm}
\end{table}
\vspace{-0.4cm}
\subsection{Object-relative State Estimation} \label{subsec:state_estimation}

The goal is to estimate the pose of the IMU ($I$), our propagation sensor, with respect to the navigation world ($W$) by measuring the relative 6-DoF poses between the navigation camera ($C$) and a set of objects of interest ($O_k$), which are provided by our DL-based pose estimator dubbed PoET. We use MaRS \cite{brommer2020mars} for multi-sensor fusion and state estimation due to it being developed with mobile robots in mind, hence being lightweight and computationally efficient. In our previous work on AI-based multi-object-relative state estimation \cite{jantos2023ai}, we introduced the concept of a multi-pose landmark sensor that benefits from concurrently processing multiple relative pose measurements during an EKF update step. We kindly refer the reader to this work for additional insights. Assuming that $N$ objects are present in the scene, the state vector $\mathbf{X}$ is given by
\vspace{-0.3cm}
\begin{align}
 \mathbf{X} = [&{\mathbf{p}}_{\scriptscriptstyle WI}^T, {\mathbf{v}}_{\scriptscriptstyle WI}^T, {\mathbf{q}}_{\scriptscriptstyle WI}^T, {\mathbf{b}}_{\scriptscriptstyle \omega}^T, {\mathbf{b}}_{\scriptscriptstyle a}^T, \\
 &{\mathbf{p}}_{\scriptscriptstyle IC}^T, {\mathbf{q}}_{\scriptscriptstyle IC}^T,{\mathbf{p}}_{\scriptscriptstyle O_0W}^T, {\mathbf{q}}_{\scriptscriptstyle O_0W}^T, \dots, {\mathbf{p}}_{\scriptscriptstyle O_NW}^T, {\mathbf{q}}_{\scriptscriptstyle O_NW}^T]^T, \nonumber
\end{align}
with the core states for state propagation being the position ${\mathbf{p}}_{\scriptscriptstyle WI}$ of the IMU, its velocity ${\mathbf{v}}_{\scriptscriptstyle WI}$ and its orientation ${\mathbf{q}}_{\scriptscriptstyle WI}$ as well as the gyroscopic bias ${\mathbf{b}}_{\scriptscriptstyle \omega}$ and the accelerometer bias ${\mathbf{b}}_{\scriptscriptstyle a}$. \textcolor{black}{We estimate the calibration between the IMU and the camera given by ${\mathbf{p}}_{\scriptscriptstyle IC}$ and ${\mathbf{q}}_{\scriptscriptstyle IC}$ . Additionally, we estimate object-worlds (${\mathbf{p}}_{\scriptscriptstyle O_kW}, {\mathbf{q}}_{\scriptscriptstyle O_kW}$) that relates the object frame to the navigation world.} The system dynamics are given by \cite{weiss2011monocular}
\begin{align}
    {\Dot{\mathbf{p}}}_{\scriptscriptstyle WI} &= {\mathbf{v}}_{\scriptscriptstyle WI}\\
    {\Dot{\mathbf{v}}}_{\scriptscriptstyle WI} &= {\mathbf{R}}_{\scriptscriptstyle WI} ~ ({\mathbf{a}}_{\scriptscriptstyle m} - {\mathbf{b}}_{\scriptscriptstyle a} - {\mathbf{n}}_{\scriptscriptstyle a}) - \mathbf{g} \\
    {\Dot{\mathbf{q}}}_{\scriptscriptstyle WI} &= \frac{1}{2} \Omega({\mathbf{\omega}}_{\scriptscriptstyle b} - {\mathbf{b}}_{\scriptscriptstyle \omega} - {\mathbf{n}}_{\scriptscriptstyle \omega}) ~ {\mathbf{q}}_{\scriptscriptstyle WI},
\end{align}
where ${\mathbf{a}}_{\scriptscriptstyle m}$ is the measured acceleration in the IMU frame, ${\mathbf{n}}_{\scriptscriptstyle a}$ is the accelerometer noise parameter, $\mathbf{g}$ is the gravity vector in $W$, ${\mathbf{\omega}}_{\scriptscriptstyle b}$ is the measured angular velocity in the IMU frame, ${\mathbf{n}}_{\scriptscriptstyle \omega}$ is the gyroscopic noise parameter, and $\Omega(\omega)$ is the quaternion multiplication matrix of $\omega$. The IMU biases are modeled as random walks.
\looseness=-1

\begin{figure}[t]
    \centering
    \includegraphics[width=0.8\columnwidth]{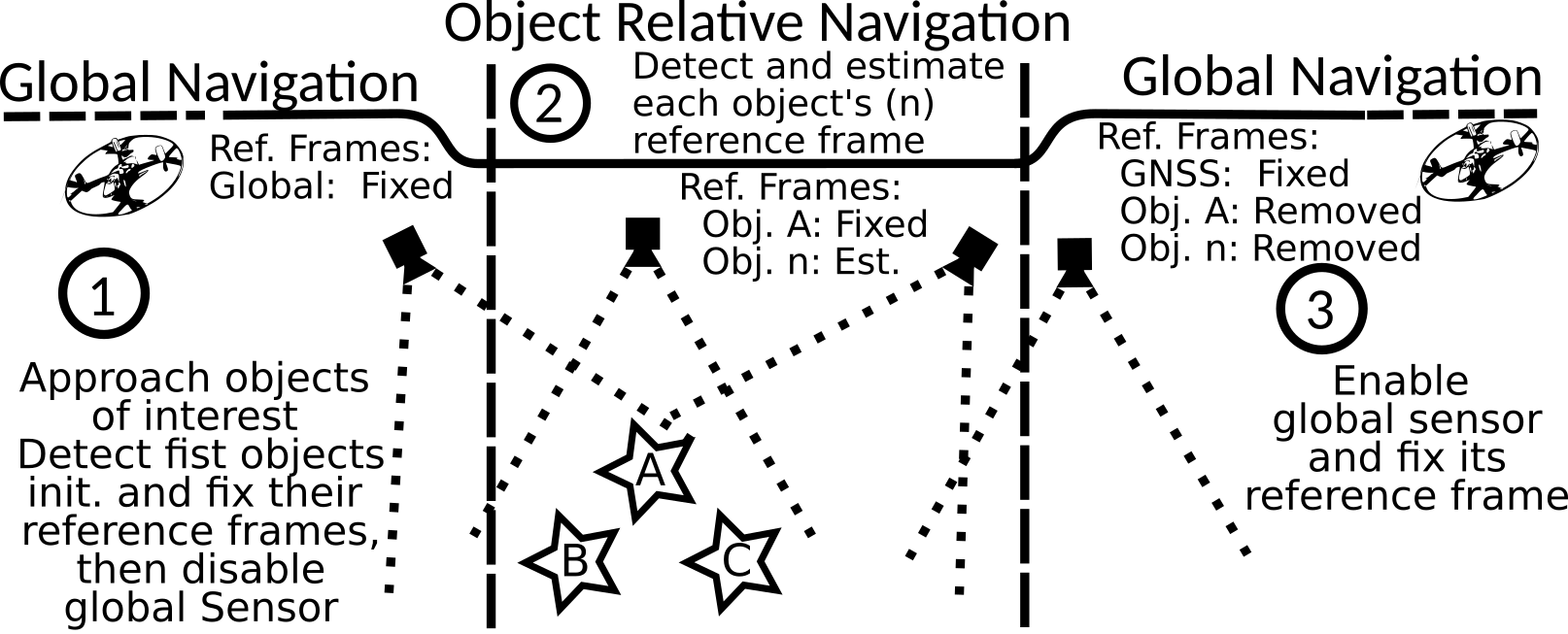}
    \caption{Schematic representation of our proposed object-relative navigation workflow. The objects of interest, e.g. infrastructure, are approached using global navigation. As soon as the objects of interest are detected, the UAV switches to object-relative navigation by estimating the states of the objects. While the global reference frame is discarded, one of the detected objects is fixed as a reference frame to render the problem observable. Once the object-relative navigation is done, the UAV switches back to global navigation and removes the object states from its estimation.}
    \label{fig:obj_relative_schematic}
    \vspace{-0.6cm}
\end{figure}

In order to perform object-relative, closed-loop UAV navigation, it is necessary that at least one object of interest is in the field of view of the UAV. Therefore, the UAV will take off and fly to a pre-defined start position using a global position sensor, for example a GNSS sensor. Once the start position is reached, the UAV hovers until it verifies that the object of interest is in its field of view. Afterwards, the global position sensor is switched off and the landmark sensor for object-relative navigation is switched on, as visualized in \cref{fig:obj_relative_schematic}. This is possible due to the modularity of MaRS that separates the propagation of the core state variables based on inertial data from state updates based on the measurements of the individual sensors. Not only does this allow for the straightforward integration of new sensors, but also allows for sensor switching while flying. When switching to the landmark sensor, the pose of an insulator in the world ($O_k$) is initialized in the filter with its first relative 6-DoF pose measurement and based on the current estimated state of the the robot:
\vspace{-0.12cm}
\begin{align}
    {\mathbf{R}}_{\scriptscriptstyle O_kW} &= {\mathbf{R}}_{\scriptscriptstyle O_kC} {\mathbf{R}}_{\scriptscriptstyle IC}^T {\mathbf{R}}_{\scriptscriptstyle WI}^T\\
    {\mathbf{p}}_{\scriptscriptstyle O_kW} &= {\mathbf{p}}_{\scriptscriptstyle O_kC} - {\mathbf{R}}_{\scriptscriptstyle O_kW} ({\mathbf{R}}_{\scriptscriptstyle WI} ~ {\mathbf{p}}_{\scriptscriptstyle IC} + {\mathbf{p}}_{\scriptscriptstyle WI})
\end{align}
It is important to highlight that for complete object-relative state estimation it is necessary to fix one of the objects as our anchor landmark $(A)$. In our previous work \cite{jantos2023ai} this was achieved by setting the Jacobian of all states associated with the anchor object $(\mathbf{p}_{O_AW}, \mathbf{R}_{O_AW})$ to zero and thus not updating the states. In this work, we follow an alternative approach, namely to perform pseudo measurements with close to zero measurement noise for the states to be fixed during each update state of the EKF. As it is possible to switch the anchor to a different landmark, e.g., in case the original anchor landmark moves out of the field of view, the current estimated object world $(\mathbf{p}_{\scriptscriptstyle O_kW}, \mathbf{R}_{\scriptscriptstyle O_kW})$ is stored as the fixed pseudo measurement $(\hat{\mathbf{p}}_{\scriptscriptstyle O_AW}, \hat{\mathbf{q}}_{\scriptscriptstyle O_AW})$. During each update step, the residual and Jacobian for every object detected is calculated and stacked according to \cite{jantos2023ai} disregarding whether the object is the anchor or not. Afterwards, the residual and Jacobian for the pseudo measurement is calculated to extend the residual and Jacobian from the regular update step. It is important to note that we perform the pseudo measurement only for the position and yaw of the anchor object pose, hence allowing the roll and pitch angles to continue to change. The reason for this is that wrongly initialized pitch and roll of the anchor object in the world will lead to an accumulation of IMU biases, which is not favourable for state estimation. The residual and Jacobian for the pseudo measurement are given by
\vspace{-0.1cm}
\begin{align}
    \tilde{z}_{\mathbf{p}_{O_{A}W}} &=  \mathbf{p}_{\scriptscriptstyle O_AW} - \hat{\mathbf{p}}_{\scriptscriptstyle O_AW}\\
    \tilde{z}_{\mathbf{R}_{O_AW}} &= 2 ~ \frac{\tilde{z}_{\mathbf{q}_{\mathbf{v},O_AW}}}{ \tilde{z}_{q_{w,O_AW}}} \\
    \tilde{z}_{\mathbf{q}_{O_AW}} &= \hat{\mathbf{q}}_{O_AW}^{-1} \otimes \mathbf{q}_{O_AW} \\
    H_{\mathbf{p}, {\mathbf{p}}_{\scriptscriptstyle O_AW}} &= \mathbf{I}_3 \\
    H_{\mathbf{R}, {\mathbf{R}}_{\scriptscriptstyle O_AW}} &= \begin{pmatrix}
0 & 0 & 0\\
0 & 0 & 0 \\
0 & 0 & 1
\end{pmatrix}
\end{align}
The remaining Jacobians for the fixed pseudo measurement with respect to the core states and other object are set to $\mathbf{0}_3$. 

After switching to the landmark sensor, the object-relative flight trajectory is calculated in the navigation world ($W$) based on the state of the anchor landmark and a pre-defined inspection mission. The waypoints are provided to the mission sequencer of the CNS Flight Stack, see \cref{fig:hw_sw_overview}. 

In contrast to our previous work \cite{jantos2023ai}, where we assumed that the objects of interest belong to different classes, in this work the insulators are identical and thus not distinguishable by class. To mitigate this issue, the landmark sensor is extended with measurement matching using a Hungarian algorithm \cite{kuhn1955hungarian} to perform a 1-to-1 matching between incoming relative 6-DoF object pose measurements and the landmarks already initialized in the filter. Given the relative object pose measurements and the current estimated robot pose, the landmark is projected into the world frame and matched to the stored landmark poses based on the distance. As the number of landmark objects is fixed beforehand in the state vector, the result of the measurement matching can be used to either discard measurements or to initialize one of the remaining object worlds in the filter. Additionally, we perform a $\chi^2$-test for each individual landmark during the update step to reject outliers within the multi-pose measurement. Even though the extrinsic parameters between the IMU and camera are part of our state vector, we do not estimate them but rather determine them during a calibration step.
\looseness=-1

\textcolor{black}{After completion of the inspection task, the UAV flies far enough from the infrastructure to switch back to a global position sensor, in order to perform the next task in the overarching mission plan.} Potential estimation errors accumulated during object-relative navigation are mitigated by reinitializing the global pose sensor with a measurement reference state that captures the offset between the current estimate and the first measurement. A schematic overview of the switching between global and object-relative navigation is presented in \cref{fig:obj_relative_schematic}.

\vspace{-0.3cm}
\section{Experiments \& Results} \label{sec:experiments}

In this section, we present the conducted real-world experiments \textcolor{black}{mimicking the challenging use case of power pole inspection in the wild} and discuss their results. First, we record data with the real system to perform offline state estimation. Second, we conduct multiple DL-based, object-relative, closed loop flights and assess the performance of our approach. The ground truth information for the state estimation evaluation is given by our MoCap system and recorded at 60 Hz.
\looseness=-1

\begin{figure}[t]
    \centering
    \includegraphics[width=1.0\columnwidth]{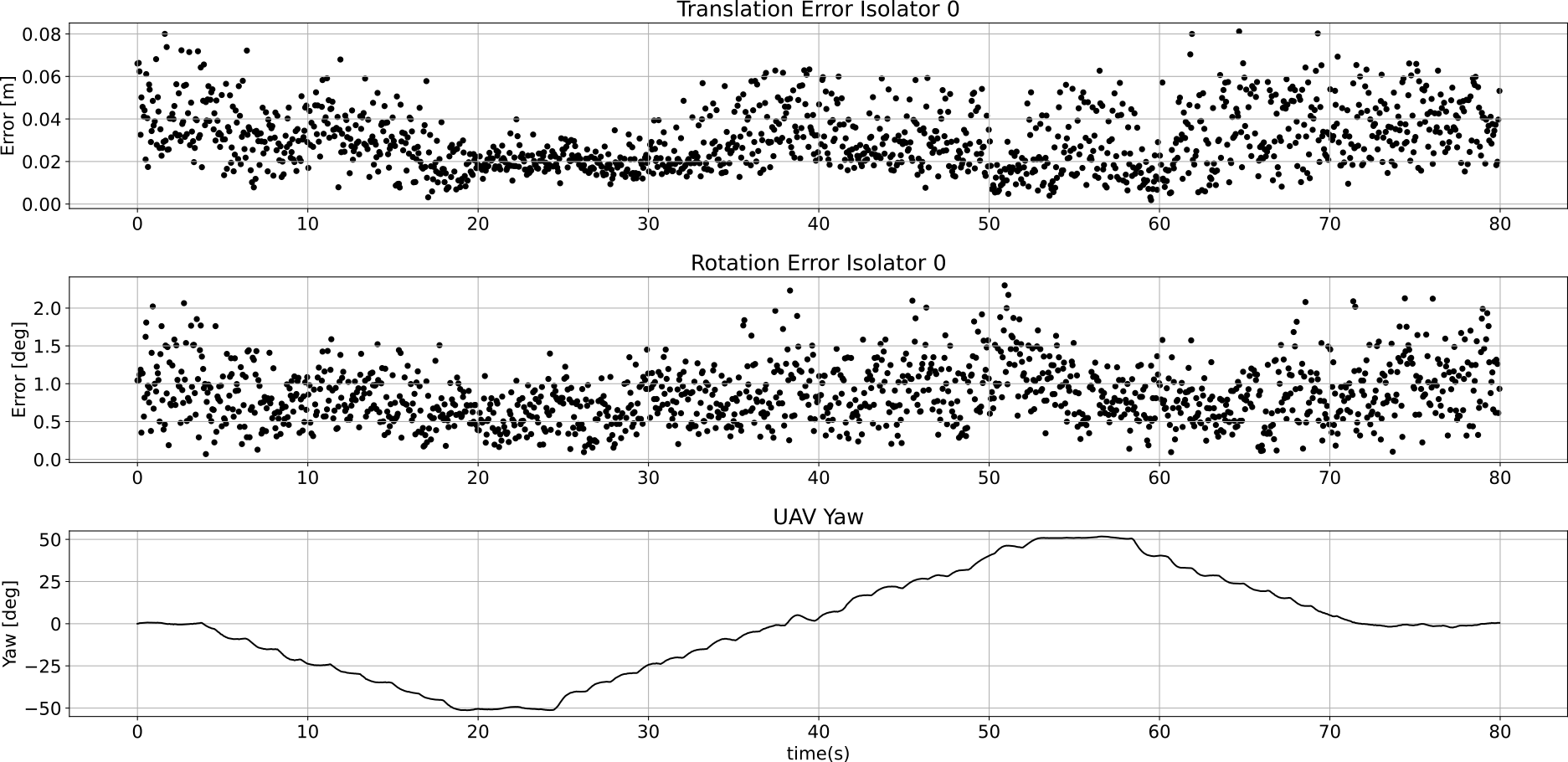}
    \vspace{-0.8cm}
    \caption{\textcolor{black}{Translation (\texttt{top}) and rotation (\texttt{middle}) error of the raw PoET measurements for the left insulator along the whole trajectory using synthetic images at 15 fps. We additionally visualize the ground truth yaw orientation of the UAV (\texttt{bottom}) to emphasize the current viewpoint.}}
    \label{fig:error_isolator}
    \vspace{-0.4cm}
\end{figure}
\begin{table}[t]
    \centering \caption{Average position/axis angle error for hovering 30s infront of the power pole for different distances/yaw angles.}
\begin{tabular}{c|ccc}
Distance/Angle & 0° & 25° & 50°\tabularnewline
\hline 
\hline 
3.0m & 0.049m / 1.15° & 0.071m / 1.76° & 0.033m / 1.49°\tabularnewline
3.3m & 0.052m / 1.74° & 0.064m / 1.64° & 0.040m / 1.29°\tabularnewline
3.6m & 0.084m / 3.45° & 0.111m / 1.86° & 0.045m / 1.47°\tabularnewline
\end{tabular}
    \label{tab:hover}
    \vspace{-0.4cm}
\end{table}
\vspace{-0.3cm}
\subsection{Object-relative State Estimation}
\label{subsec:offline}

\begin{figure*}[t]
    \centering
    \includegraphics[width=1.0\columnwidth]{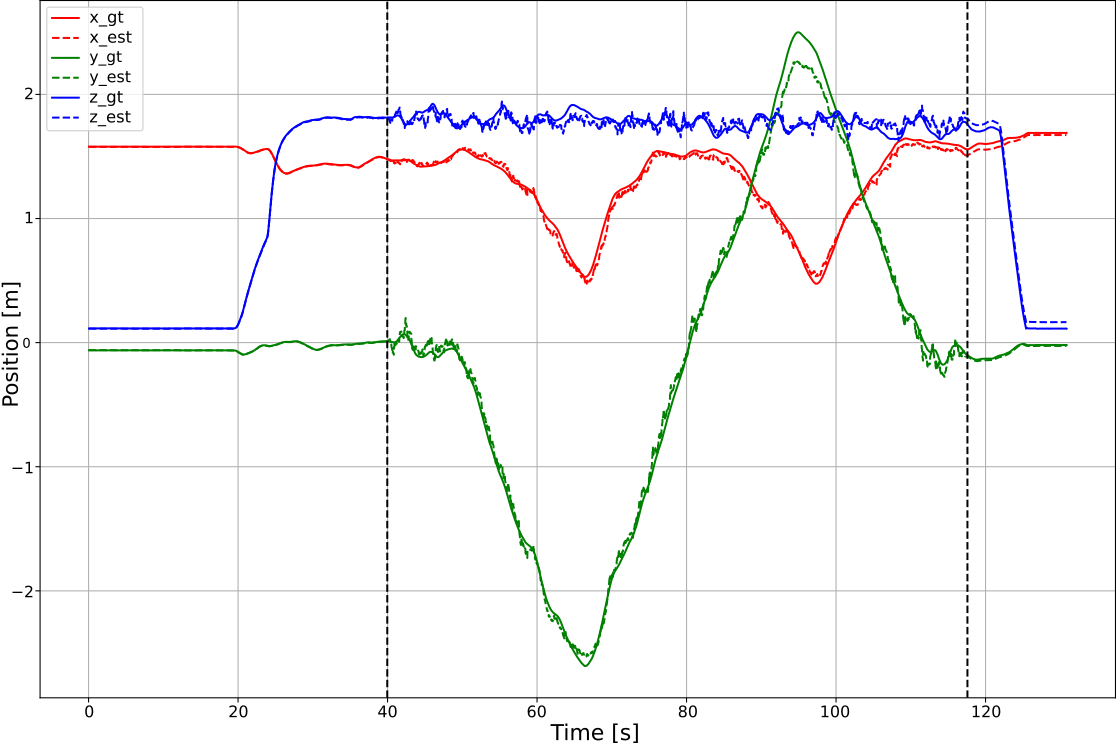}
    \includegraphics[width=1.0\columnwidth]{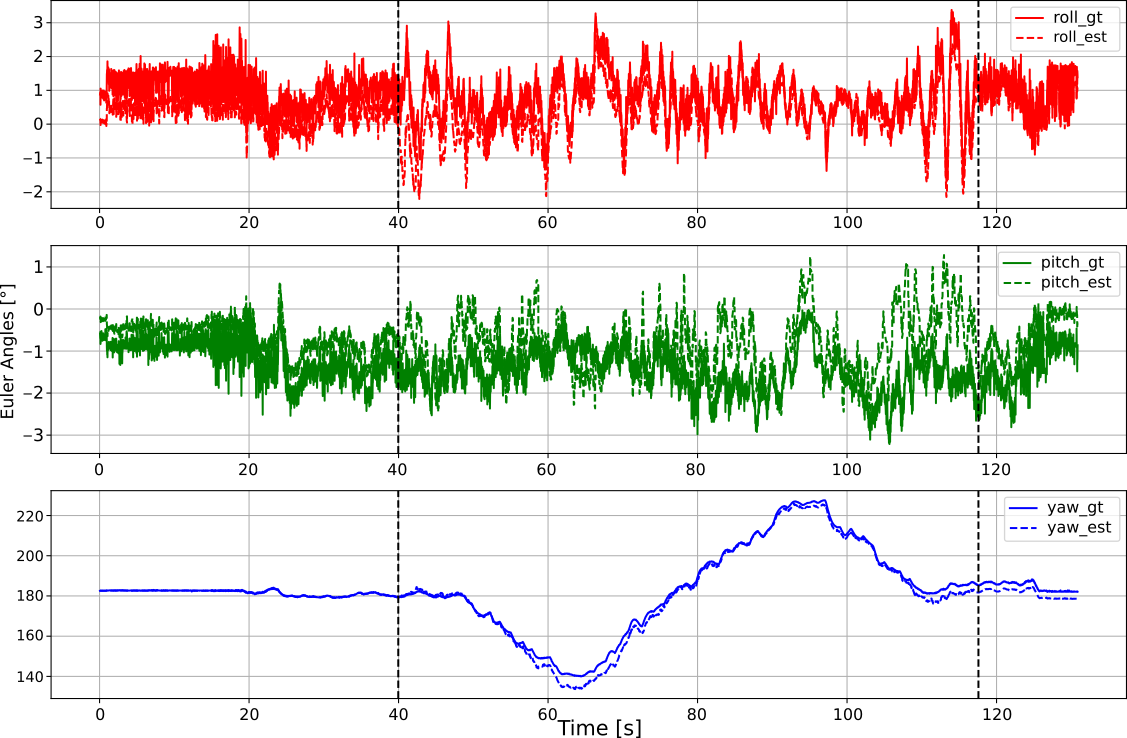}
    \vspace{-0.4cm}
    \caption{Comparison between the state estimate (\texttt{mars}) and the ground truth (\texttt{gt}) for position (\texttt{left}) and orientation (\texttt{right}) for a closed-loop flight. Note, the black dotted line indicates the exact time switching between global navigation and object-relative navigation. We can observe that the estimator is able to accurately track its state.}
    \label{fig:example_trajectory}
    \vspace{-0.5cm}
\end{figure*}

In order to solely evaluate the performance of our state estimator, avoiding undesirable behavior and corrections caused through wrong estimates in closed-loop flight, we first conducted two MoCap-based experiments with the real system introduced in \cref{subsec:hw_sw_setup} and performed offline sensor fusion and state estimation. For both experiments, we report the root mean squared error (\texttt{RMSE}) for the position and the orientation as well as the maximum position error (\texttt{PE}). First, we let the UAV hover for 30 seconds at certain distances and under specific yaw angles with respect to the power pole. The idea is to look into the effect of the viewpoint on the performance of the state estimator. As presented in \cref{tab:hover}, the RMSE for position and orientation increases with distance to the power pole for all viewing angles. On the other hand, the results for different viewing angles indicate that the appearance of the power pole in the image has an effect on the consistency of the predictions from PoET. A possible explanation is that with an increased viewing angle the baseline formed by the power pole is larger and thus benefiting the depth estimation and the pose estimate in general.

Second, as depicted in \cref{fig:title_picture}, we define an arc trajectory around the power pole, simulating an inspection flight. The UAV takes off, hovers for about 10 seconds and then flies around the power pole on an arc at fixed distance $d$ and \mbox{height $h$} until it reaches $\pm50^\circ$ in yaw from its starting orientation, while keeping the power pole centered in the navigation camera. We would like to point out that even though the exact trajectory performed would not be possible in a real-world scenario due to the presence of live power lines, it can be easily adapted to meet these constraints. The focus of the current work lies on evaluating the performance of our proposed approach with a simple and reproducible trajectory. To that end, we repeated this flight seven times with distance $d=3.3$~m, height $h=2.0$~m, and recorded sensor data for offline evaluation as well as MoCap data for ground truth comparison. It is important to note, that our DL-based pose estimator was running onboard of the UAV and we recorded its predictions as well. Additionally, we utilized ground truth data to simulate the recorded trajectories in IsaacSim to generate images to investigate the influence of the Sim2Real transfer on the estimation performance. After generating the images, we processed them with PoET on the UAV onboard computer and stored the estimated relative 6-DoF object poses. These synthetically generated pose measurements were fused with the collected real-world IMU data as a comparison experiment. 

Based on the results from the homography comparison, see \cref{tab:torch_trt_homography}, and the hover experiment, see \cref{tab:hover}, we set the measurement noise to 10~cm and $5^\circ$ for all three axis of position and rotation respectively. In \cref{fig:error_isolator}, we plot the translation and rotation error given the raw, DL-based pose measurements for the left insulator for one of the IsaacSim trajectories. We can see that the position measurements are quite noisy across the complete trajectory, except for the period where the UAV is close to the insulator, around image 400. Especially when hovering in front of the power pole, the raw position measurement can be up to 8~cm off for the simulated data. Depending on the moment of switching to object-relative navigation, the object could be wrongly initialized in the filter thus substantially impacting the estimation performance. Therefore, and due to possible errors introduced by Sim2Real, we chose the measurement noise slightly higher than the observed maximum errors. While the left and right insulator will be at one point of the trajectory closer to the UAV, the insulator at the top of the power pole is at the center of the arc, has a constant distance to the UAV and thus a more consistent error. Hence, we chose the top insulator as our fixed object in the EKF for all experiments. Even though switching the anchor landmark is possible, we kept the top insulator as our fixed landmark throughout the whole trajectory.
\begin{table}[t]
    \vspace{-0.5cm}
    \centering \caption{Results for object-relative state estimation using recorded IMU data and either real or synthetic images as input to our DL-based object pose estimator. We report the STD for the RMSE metric.}
\scalebox{0.74}{
\begin{tabular}{c|c|c|c|c|c|c}
\multirow{2}{*}{Flight} & \multicolumn{3}{c|}{Real World Images} & \multicolumn{3}{c}{Synthetic Images}\tabularnewline
\cline{2-7} \cline{3-7} \cline{4-7} \cline{5-7} \cline{6-7} \cline{7-7} 
 & RMSE {[}m{]} & RMSE {[}°{]} & Max PE {[}m{]} & RMSE {[}m{]} & RMSE {[}°{]} & Max PE {[}m{]}\tabularnewline
\hline 
\hline 
1 & 0.151 \color{black} $\pm$ 0.06& 5.85 \color{black} $\pm$ 1.5& 0.303 & 0.135 \color{black} $\pm$ 0.05& 4.23 \color{black} $\pm$ 1.0& 0.291\tabularnewline
2 & 0.170 \color{black} $\pm$ 0.09& 4.13 \color{black} $\pm$ 1.7& 0.407 & 0.208 \color{black} $\pm$ 0.07& 5.48 \color{black} $\pm$ 0.6& 0.464\tabularnewline
3 & 0.200 \color{black} $\pm$ 0.08& 3.66 \color{black} $\pm$ 1.6& 0.359 & 0.086 \color{black} $\pm$ 0.04& 2.23 \color{black} $\pm$ 0.6& 0.181\tabularnewline
4 & 0.173 \color{black} $\pm$ 0.05& 2.57 \color{black} $\pm$ 1.1 & 0.299 & 0.247 \color{black} $\pm$ 0.14& 6.38 \color{black} $\pm$ 3.0& 0.523\tabularnewline
5 & 0.195 \color{black} $\pm$ 0.07& 3.49 \color{black} $\pm$ 1.4& 0.317 & 0.134 \color{black} $\pm$ 0.06& 4.64 \color{black} $\pm$ 2.2& 0.281\tabularnewline
6 & 0.136 \color{black} $\pm$ 0.06& 3.20 \color{black} $\pm$ 1.5& 0.319 & 0.272 \color{black} $\pm$ 0.15& 8.67 \color{black} $\pm$ 4.3& 0.587\tabularnewline
7 & 0.162 \color{black} $\pm$ 0.07& 2.98 \color{black} $\pm$ 1.2& 0.335 & 0.106 \color{black} $\pm$ 0.04& 2.22 \color{black} $\pm$ 0.8& 0.220\tabularnewline
\hline 
Mean & 0.170 \color{black} $\pm$ 0.07 & 3.70 \color{black} $\pm$ 1.4& 0.334 & 0.170 \color{black} $\pm$ 0.08& 4.84 \color{black} $\pm$ 1.8& 0.364\tabularnewline
\hline 
\end{tabular}
}
    \label{tab:trajectory_offline}
    \vspace{-0.3cm}
\end{table}
In \cref{tab:trajectory_offline} the results for offline trajectory evaluation are presented. For evaluation, we only consider the segments during which the UAV performed object-relative state estimation, namely after the end of the hover phase. Our approach achieves the same mean RMSE for the position for the real world images as well as the simulated images. In terms of the mean RMSE for orientation and mean maximum position error, a similar performance is achieved across the seven recorded flights. However, we can observe that for individual flights the performance differs when using real-world or synthetic images. 
Overall, our state estimator achieves a lower orientation RMSE and maximum position error with real images. The reason for that might be due to time synchronization issues between the IMU and the simulated images. Nonetheless, these results confirm that the Sim2Real transfer in combination with the homography between two different camera matrices is working for object-relative state estimation.
\vspace{-0.3cm}
\subsection{Closed-loop Navigation}

The previous experiments highlight that relative pose measurements from a DL-based 6-DoF object pose estimator, solely trained on synthetic data, can be used for object-relative state estimation by fusing them with IMU data. With the aim of showing that our system is capable of object-relative navigation, we conducted ten closed-loop flights with onboard processing and present the results in \cref{tab:ai_closed_loop}. We considered the same trajectory as described in \cref{subsec:offline} and implemented the waypoint calculation and sensor switching strategy described in \cref{subsec:state_estimation}. Based on the current estimated state of the anchor landmark $(\mathbf{p}_{\scriptscriptstyle O_AW}, \mathbf{R}_{\scriptscriptstyle O_AW})$ and UAV $(\mathbf{p}_{\scriptscriptstyle WI}, \mathbf{R}_{\scriptscriptstyle WI})$ the trajectory waypoints were calculated such that the distance was kept fixed and the z-component corresponded to the UAV's current height. Similarly to the offline evaluation, we only considered the object-relative part of the flight for the metric calculation. Additionally, we increased the measurement noise to 20~cm and $10^\circ$ for position and orientation respectively to account for possible errors introduced through flight dynamics. We observe a similar performance across the closed loop flights when compared to the offline evaluation. However, the estimator is more prone to having a higher maximum position error when performing closed-loop navigation. In \cref{fig:example_trajectory}, we compare the estimated state of the UAV to the ground truth for position and orientation. As we can see, the UAV was able to correctly estimate its state for the entirety of the trajectory, except for a small segment where the power pole was viewed from the side. Our experiments show that we can conduct closed-loop flights and reliably reproduce the state estimation performance across multiple runs. An example closed-loop flight is presented in the supplementary video.

\begin{table}[t]
    \centering \caption{Results for closed-loop, object-relative navigation. We report the STD for the RMSE metric.}
\scalebox{0.9}{
\begin{tabular}{c|c|c|c}
Flight & RMSE {[}m{]} & RMSE {[}°{]} & Max PE {[}m{]}\tabularnewline
\hline 
\hline 
1 & 0.127 \color{black} $\pm$ 0.06& 2.79 \color{black} $\pm$ 1.3& 0.319\tabularnewline
2 & 0.147 \color{black} $\pm$ 0.05& 3.34 \color{black} $\pm$ 1.6& 0.300\tabularnewline
3 & 0.123 \color{black} $\pm$ 0.06& 2.72 \color{black} $\pm$ 1.4& 0.328\tabularnewline
4 & 0.110 \color{black} $\pm$ 0.05& 3.17 \color{black} $\pm$ 1.6& 0.264\tabularnewline
5 & 0.153 \color{black} $\pm$ 0.07& 3.49 \color{black} $\pm$ 1.4& 0.425\tabularnewline
6 & 0.131 \color{black} $\pm$ 0.05& 2.70 \color{black} $\pm$ 1.3& 0.464\tabularnewline
7 & 0.140 \color{black} $\pm$ 0.08& 2.53 \color{black} $\pm$ 1.2& 0.448\tabularnewline
8 & 0.138 \color{black} $\pm$ 0.06& 2.61 \color{black} $\pm$ 1.1& 0.272\tabularnewline
9 & 0.149 \color{black} $\pm$ 0.08& 2.65 \color{black} $\pm$ 1.2& 0.358\tabularnewline
10 & 0.136 \color{black} $\pm$ 0.07& 3.54 \color{black} $\pm$ 1.4& 0.410\tabularnewline
\hline 
Mean & 0.135 \color{black} $\pm$ 0.06& 2.95 \color{black} $\pm$ 1.3& 0.359\tabularnewline
\hline 
\end{tabular}
}
    \label{tab:ai_closed_loop}
    \vspace{-0.4cm}
\end{table}
\vspace{-0.3cm}
\section{Conclusion} \label{sec:concluison}

In this letter, we presented a real-time capable UAV system for semantic, closed-loop, object-relative navigation with onboard processing of all information and a minimal sensor configuration. Semantic navigation is essential for tasks like object following or infrastructure inspection. Object-relative localization is achieved by fusing DL-based, relative 6-DoF object pose measurements with IMU data\textcolor{black}{, not requiring any additional sensors}. \textcolor{black}{However, it relies on the objects of interest being in the field of view of the navigation camera and thus requiring a safety system.} Training such DL-based methods is data intensive as it requires large sets of precisely annotated images. We showcased that a DL-based network solely trained on synthetic images and made camera-agnostic through mapping between different camera parameters achieves satisfying performance on real-world data. In combination with the minimal sensor configuration, our proposed approach is quickly adaptable to a variety of tasks and mobile robot platforms. In order to realize a real-time capable, autonomous system, we optimized our DL-based 6-DoF pose estimator for the target hardware and implemented a sensor switching and waypoint calculation mechanism taking into account the current estimate. To validate our system, we conducted multiple real-world experiments that represent infrastructure inspection flights for power poles. \textcolor{black}{In future work, we will conduct a robustness analysis for increased flight speed regarding closed-loop behavior, state estimation and object pose estimation.}
\looseness=-1
\vspace{-0.3cm}
\bibliographystyle{IEEEtran.bst}
\bibliography{IEEEabrv, root.bib}
\end{document}